%% file: main.tex
\documentclass[10pt,twocolumn,letterpaper]{article}

\usepackage[pagenumbers]{iccv} 

\usepackage{float}
\usepackage{multirow}
\usepackage{booktabs}
\PassOptionsToPackage{table,xcdraw}{xcolor}
\usepackage{colortbl}
\usepackage{graphicx}
\usepackage{makecell} 
\usepackage{threeparttable}
\usepackage[accsupp]{axessibility}  


\definecolor{iccvblue}{rgb}{0.21,0.49,0.74}
\usepackage[pagebackref,breaklinks,colorlinks,allcolors=iccvblue]{hyperref}

\def\confName{ICCV}
\def\confYear{2025}

\title{Unveiling the Invisible: Reasoning Complex Occlusions Amodally with AURA}

\author{
Zhixuan Li$^{1}$ \quad
Hyunse Yoon$^2$ \quad
Sanghoon Lee$^2$ \quad 
Weisi Lin$^{1*}$ \quad
\\
    $^1$College of Computing and Data Science, Nanyang Technological University, Singapore \\
    $^2$Department of Electrical and Electronic Engineering, Yonsei University, Korea \\ 
{\tt\small zhixuanli520@gmail.com, {hsyoon97, slee}@yonsei.ac.kr, wslin@ntu.edu.sg}
}

\begin{document}

\twocolumn[{
\renewcommand\twocolumn[1][]{#1}
\maketitle
\begin{center}
    \centering
    \includegraphics[width=1\linewidth]{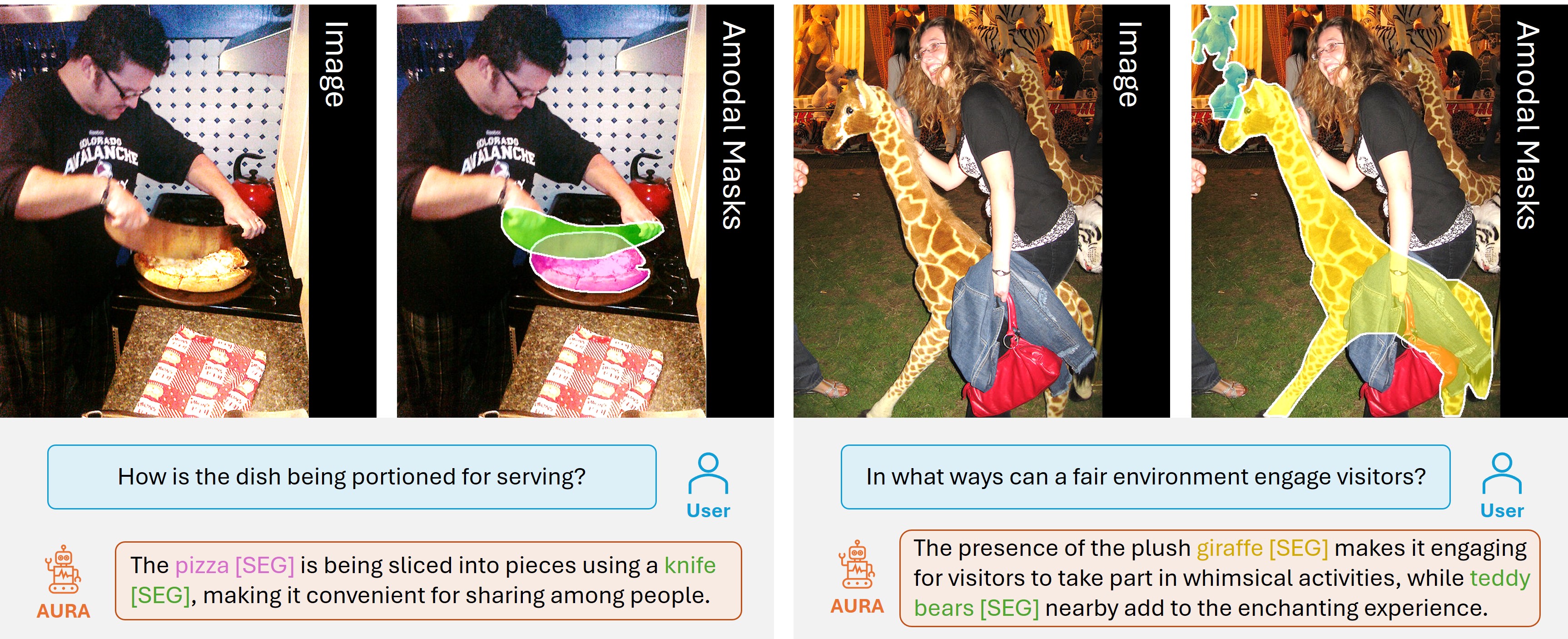}
    \captionof{figure}{We introduce AURA, a multi-modal approach designed for reasoning the amodal segmentation mask including both visible and occlusion regions based on the user's question. AURA can deduce the implicit purpose underneath the question and respond with the textual answer along with predicted amodal masks for various objects. Best viewed in color.}
    \label{fig:motivation}
\end{center}
}]

\def\thefootnote{*}\footnotetext{Corresponding author.}

\input{sec/0_abstract}    
\input{sec/1_main_text}

\maketitlesupplementary

In the supplementary material, we introduce the following contents, including:

\begin{itemize}
    \item Generalization ability is demonstrated in Section \ref{sec:suppl_generalization},
    \item Experimental results in Section \ref{sec:suppl_exp},
    \item Discussions in Section \ref{sec:suppl_discussions},
    \item Additional Information of the Dataset in Section \ref{sec:suppl_dataset}.
\end{itemize}


\section{Generalization on Cloud Occluded Region Segmentation}
\label{sec:suppl_generalization}

We showcase the proposed method's potential of generalizing to the meteorological image, which is included to qualitatively demonstrate the potential of our method to generalize across domains. The example is illustrative with no quantitative results, and is not part of the core evaluation presented in the main paper.

To be specific, cloud cover is a major challenge in remote sensing and meteorological observations, as it obscures ground-level features that are critical for weather modeling, environmental monitoring, and climate studies. Accurate retrieval of ground information under cloud cover is essential for enhancing the quality of meteorological data and improving predictions.

Ground-level observations, such as vegetation health, urban heat distribution, and soil moisture, play a vital role in meteorological models. However, cloud coverage often introduces uncertainties in these observations, affecting the accuracy of derived climate indicators. Besides, meteorological models rely on ground information for initial boundary conditions, such as vegetation cover for evapotranspiration modeling or urban features for heat island studies. Addressing cloud-induced occlusions can significantly improve the reliability of such models. Furthermore, during extreme weather events like tropical cyclones or heavy rainfall, cloud cover is prevalent. Recovering occluded ground-level features aids in post-disaster assessment and supports real-time meteorological decision-making. 

It is worth noticing that the dynamic nature of cloud cover, influenced by meteorological parameters such as wind speed, humidity, and atmospheric pressure, makes it critical to develop robust methods that can adaptively infer ground information under varying cloud conditions.

We evaluate the proposed AURA method for reconstructing ground-level information obscured by cloud cover using amodal segmentation, enabling meteorologists to retrieve critical land-surface data under cloudy conditions.

\begin{figure*}[t]
    \centering
    \includegraphics[width=0.78\linewidth]{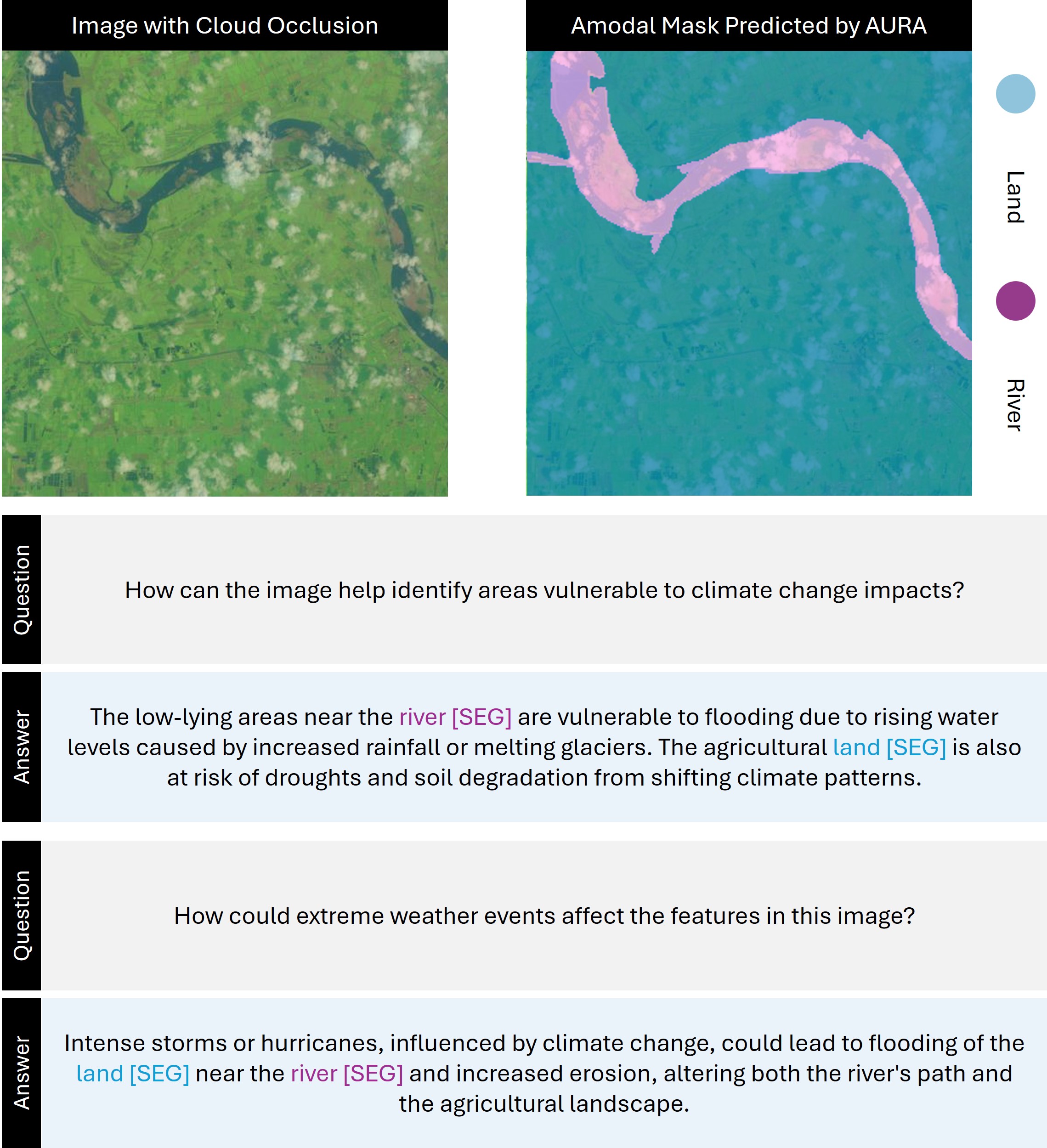}
    \caption{The qualitative result for demonstrating the generalization ability of AURA on real-world scenarios such as climate change analysis. Best viewed in color.}
    \label{fig:suppl_exp_cloud_seg_reasoning}
\end{figure*}

As shown in Figure~\ref{fig:suppl_exp_cloud_seg_reasoning}, the effectiveness of AURA is evaluated using a cloud-occluded image from the RICE dataset~\cite{lin2019remote}. This dataset is derived from the Landsat 8 OLI/TIRS dataset~\cite{roy2014landsat}, which comprises paired images captured under cloud-occluded and non-occluded conditions. The Landsat 8 OLI/TIRS dataset, developed by NASA and the United States Geological Survey (USGS), offers high-resolution multispectral and thermal imagery to support global environmental monitoring, climate change research, and sustainable resource management. For this study, we annotated ten images from the RICE dataset with ground-truth amodal masks and question-answer pairs. AURA was fine-tuned on eight images, reserving the remaining two for validation. The same model structure and parameter settings of AURA are used, as described in Section 5 of the main paper.

Figure~\ref{fig:suppl_exp_cloud_seg_reasoning} presents a qualitative result on the validation set. The results highlight AURA's capability to segment complete shapes despite cloud occlusions, demonstrating its potential to comprehend ground-level conditions based on meteorological knowledge.

\section{Experiments}
\label{sec:suppl_exp}

\subsection{Experimental Settings}

All of the experiments are conducted on the proposed AmodalReasonSeg dataset. No extra datasets other than the proposed AmodalReasonSeg dataset are used to ensure fairness in comparison.
There are no other datasets used in our experiments. The pre-trained weights of LLaVA-7B-v1-1 and SAM-Vit-H are used as the initialized weights of the multi-modal LLM as well as the vision backbone and mask decoders, respectively.

\subsection{Training and Inference Efficiency}

Regarding training time, LISA requires 8 hours for either visible or amodal segmentation, whereas AURA takes only 10 hours to handle both. 

Under the same testing environment, LISA achieves 0.33 fps, whereas AURA achieves 0.28 fps.

\subsection{Evaluation Details}

\textbf{Evaluation of amodal segmentation methods.} 
The evaluation of these models is performed by computing metrics between all predicted objects and the target object, as these models are designed to predict all objects in the scene and cannot identify which specific object should be predicted. 

\noindent \textbf{Explanation of the evaluation of MLLM-based methods.} 
First, MLLM methods like LISA and LLM-Seg predict one object at a time and are evaluated per prediction–target match, ignoring any unmatched targets. Because most images contain multiple objects, these methods tend to perform suboptimally.
Second, for MLLM-based methods that can predict multiple objects simultaneously, such as GVSA, OMG-LLaVA, GLaMM, and PSALM, the evaluation protocol is consistent with that used for AURA.


\section{Discussions}
\label{sec:suppl_discussions}

\subsection{Difference between Reasoning Segmentation and Referring Segmentation}

The key difference between the reasoning segmentation and the referring segmentation tasks lies in the content of the questions and answers.

In the reasoning segmentation task, questions are abstract and do not explicitly reference specific objects. The answers include both textual elaborations and segmentation masks for the objects mentioned. For example, the question, ``How can I accessorize my kitchen for cooking efficiently?" requires understanding and reasoning without direct object references. The corresponding answer might be, ``Utilize the bottles[SEG] on the countertop for easy access to oils or spices, and rely on the sink[SEG] for rinsing utensils and ingredients during food preparation." This answer provides both the segmentation masks (denoted as ``[SEG]") and explanatory details to address the question.

In contrast, the referring segmentation task involves questions that explicitly mention the objects to be segmented. For instance, the question, ``Where is the orange? Please segment it.", explicitly identifies the object of interest. The typical answer is concise, such as ``It is [SEG]", and lacks additional elaborations.

\subsection{Difference between Amodal Reasoning Segmentation and other Amodal Segmentation Tasks}

The proposed amodal reasoning segmentation task differs from existing amodal segmentation tasks by enabling interaction with users through textual input questions and providing elaborative textual answers.

While traditional amodal segmentation tasks—spanning semantic, instance, and panoptic levels—are limited to processing only image or video inputs, the proposed task introduces the ability to accept textual questions as input, infer the implicit purpose underneath these questions, and produce textual answers alongside segmentation masks. The ability of language interaction significantly broadens the potential for real-world applications.

\subsection{Applicability of Existing Reasoning Segmentation Method on the Proposed New Task}

There are various existing reasoning segmentation methods proposed, like LISA. These methods were originally designed for segmenting visible masks, and can be applied to the proposed new task - Amodal Reasoning Segmentation by training with ground-truth question-answer pairs and amodal segmentation masks. 

However, methods like LISA need two major improvements for this task. First, LISA can only predict visible or amodal masks, requiring an additional decoder to handle both simultaneously. Second, LISA predicts only one object, while AURA handles multiple objects in a single inference by modifying [SEG] tokens. With these changes, LISA can only be upgraded to the “Baseline” method in Table 1 of our main paper. 

Then the task-specific designs, including the proposed Occlusion Condition Encoder and the Spatial Occlusion Encoder, are required to handle the occlusion problem, which is the core difference between the traditional visible segmentation task and the focused amodal segmentation task in the paper.


\begin{figure}[t]
    \centering
    \includegraphics[width=1\linewidth]{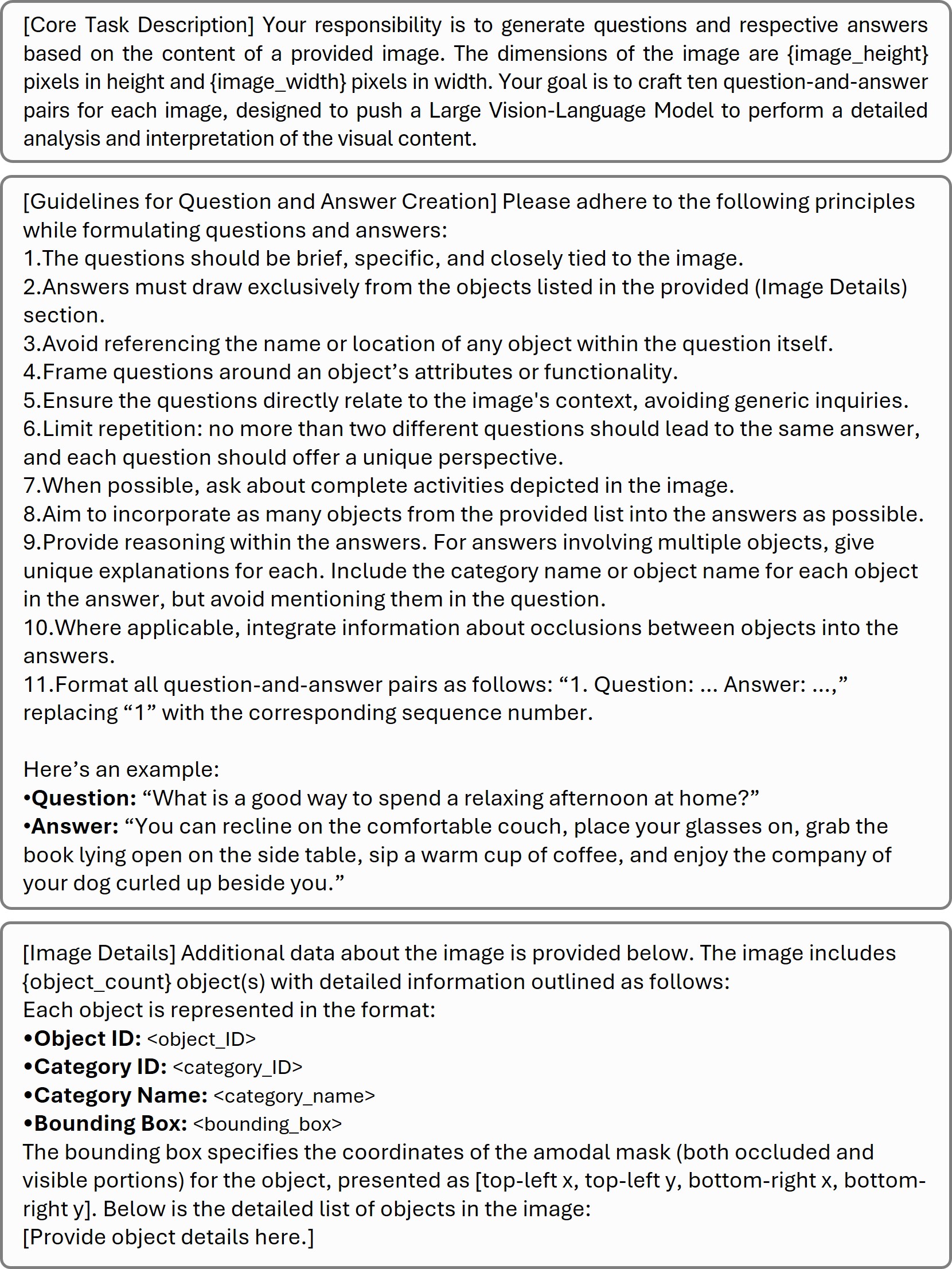}
    \caption{The designed prompt template used for guiding the GPT, including the core task description for the ChatGPT-4o, the guidelines for question and answer creation, and the details about the image and objects in it.}
    \label{fig:prompt_template}
\end{figure}

\section{Dataset}
\label{sec:suppl_dataset}

\subsection{Exhibition of Various Perspectives}

\begin{figure*}[htbp]
    \centering
    \includegraphics[width=0.99\linewidth]{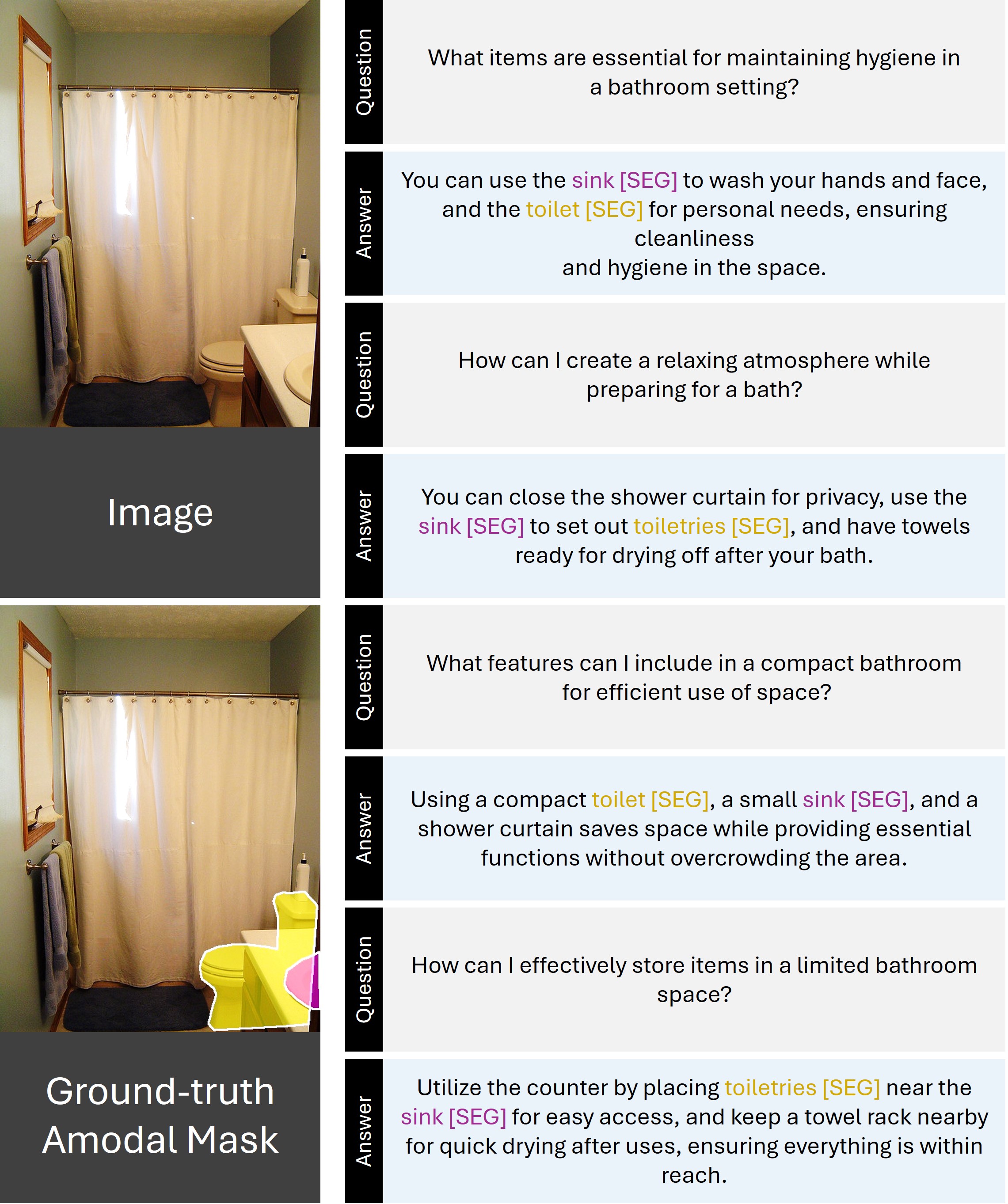}
    \caption{The first example of the proposed AmodalReasonSeg dataset shows various questions and answers from diverse perspectives for the same image. Best viewed in color.}
    \label{fig:suppl_dataset_questions_2}
\end{figure*}

\begin{figure*}[htbp]
    \centering
    \includegraphics[width=0.99\linewidth]{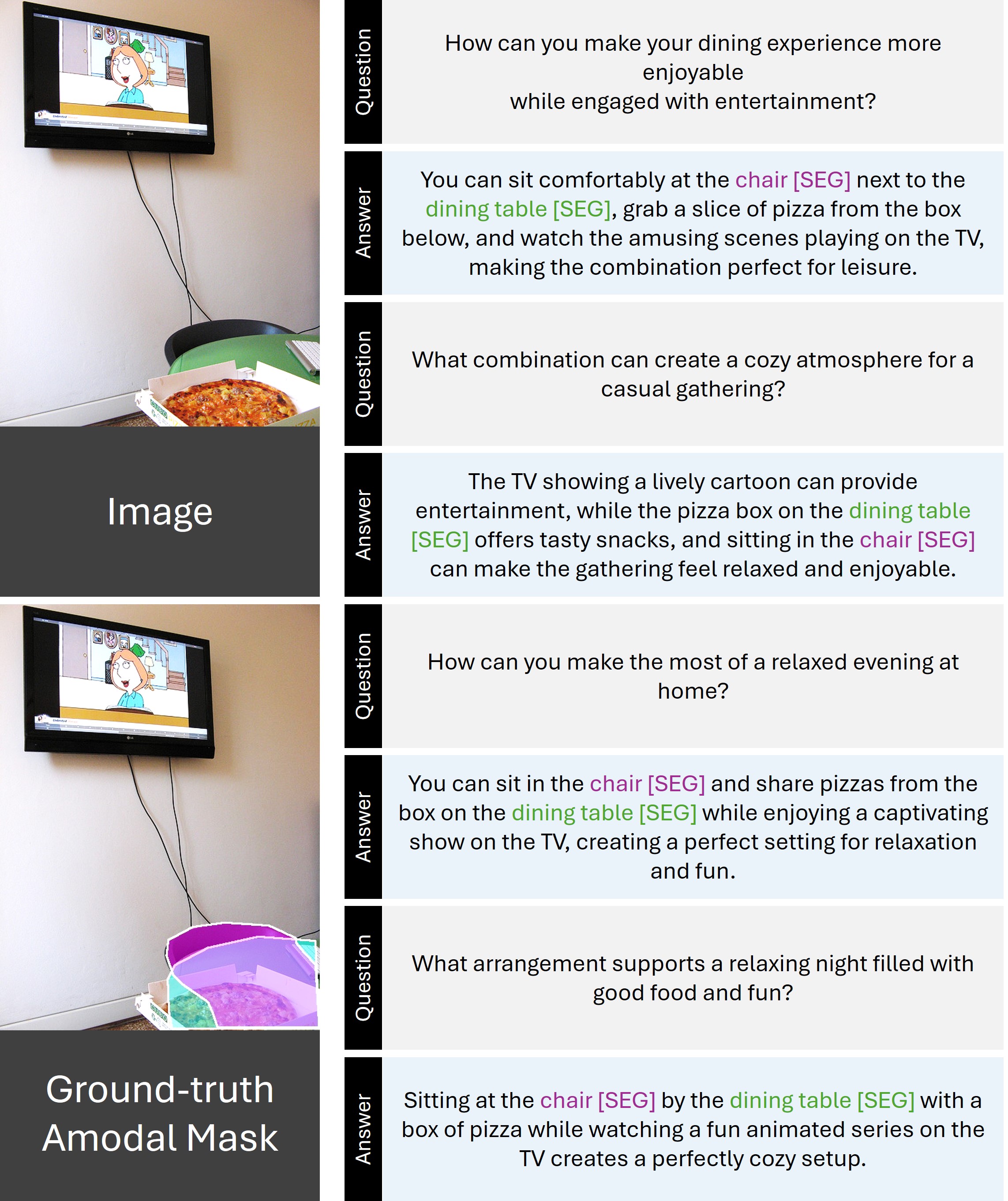}
    \caption{The second example of the proposed AmodalReasonSeg dataset shows various questions and answers from diverse perspectives for the same image. Best viewed in color.}
    \label{fig:suppl_dataset_questions_1}
\end{figure*}

In the created dataset AmodalReasonSeg, there are 11.3 pairs of questions and answers annotated for each image on average. To ensure the diversity of the question-and-answer pairs provided in this dataset, we specifically add a requirement in the prompt template used for guiding ChatGPT-4o to generate question-and-answer pairs from different perspectives to cover various potentials. As shown in Figures \ref{fig:suppl_dataset_questions_2} and \ref{fig:suppl_dataset_questions_1}, two cases of the proposed AmodalReasonSeg dataset are shown to exhibit the diversity of question-and-answer pairs in this dataset.

\subsection{Prompt Template for GPT}

We present the designed prompt template in Figure~\ref{fig:prompt_template}. This template specifies detailed requirements and provides comprehensive information about the image and objects, ensuring the generation of high-quality questions and answers by ChatGPT-4o. Additionally, it includes an example to clarify the requirements and assist ChatGPT-4o in understanding them effectively.

\subsection{Average count of question-answer pairs per image.} 

The dataset contains an average of 11.3 QA pairs per image. Initially, GPT-4o generated 10 QA pairs per image, which human annotators subsequently revised and occasionally supplemented during verification to enhance data quality, resulting in a final average above 10.

{
    \small
    \bibliographystyle{ieeenat_fullname}
    \bibliography{main}
}

\end{document}

%% file: sec/0_abstract.tex
\begin{abstract}

Amodal segmentation aims to infer the complete shape of occluded objects, even when the occluded region's appearance is unavailable. However, current amodal segmentation methods lack the capability to interact with users through text input and struggle to understand or reason about implicit and complex purposes. While methods like LISA integrate multi-modal large language models (LLMs) with segmentation for reasoning tasks, they are limited to predicting only visible object regions and face challenges in handling complex occlusion scenarios. To address these limitations, we propose a novel task named amodal reasoning segmentation, aiming to predict the complete amodal shape of occluded objects while providing answers with elaborations based on user text input. We develop a generalizable dataset generation pipeline and introduce a new dataset focusing on daily life scenarios, encompassing diverse real-world occlusions. Furthermore, we present AURA (Amodal Understanding and Reasoning Assistant), a novel model with advanced global and spatial-level designs specifically tailored to handle complex occlusions. Extensive experiments validate AURA's effectiveness on the proposed dataset. The code, model, and dataset are released on this \href{https://zhixuanli.github.io/projects/li2025aura/index.html}{page}.

\end{abstract}

%% file: sec/1_main_text.tex
\section{Introduction}
\label{sec:intro}

With the development of the Segment Anything Model (SAM) series methods~\cite{kirillov2023segment,xiong2024efficientsam,ke2024segment}, predicting the segmentation mask of visible regions has been largely solved. Amodal segmentation aims to predict the entire shape of an object, including both visible and occluded regions, which is naturally challenging since visual cues for the occluded regions are unavailable. Predicting the intact shape of occluded objects is crucial for various real-world applications, such as autonomous driving~\cite{qi2019amodal,siam2017deep,geiger2012we}, robotics~\cite{back2022unseen,cheang2022learning,jiang2025exploring}, and image editing~\cite{xu2024amodal,zhan2020self}.

Existing amodal segmentation approaches have covered the fields including semantic~\cite{breitenstein2022joint,sekkat2024synthdrive}, instance~\cite{li2016amodal,zhu2017semantic,liu2024blade}, and panoptic~\cite{mohan2022amodal} segmentation. Although existing amodal segmentation methods~\cite{hu2019sail,qi2019amodal,follmann2019learning,xiao2021amodal} can address the problem of inferring occluded object shapes to some extent, they lack the ability to integrate with large language models (LLMs)~\cite{touvron2023llama,zhang2023llamaadapter,touvron2023llamav2}, especially in performing reasoning and segmentation based on scene context, limiting their ability in interacting with users and understanding the implicit purpose underneath users questions.
Meanwhile, although existing reasoning segmentation methods~\cite{lai2024lisa,wang2024llm} can work well on predicting the visible segmentation, they cannot predict the complete amodal segmentation mask, limiting their ability to reason in scenarios with complex occlusions.
As a matter of fact, there is currently a significant gap in the availability of datasets designed for amodal reasoning segmentation, as well as in the development of algorithms specifically tailored to this task.

To address this gap, we propose a novel task named Amodal Reasoning Segmentation (ARS). As shown in Figure \ref{fig:motivation}, the ARS task aims to interact with the user through textual questions, reasoning in the complex scene, and predicting accurate segmentation masks with textual answers. 
For this task, we present a highly generalizable data collection pipeline and construct a new dataset, AmodalReasonSeg, which covers daily life scenes (encompassing both indoor and outdoor environments) derived from the challenging COCOA-cls dataset~\cite{zhu2017semantic,follmann2019learning}. The AmodalReasonSeg dataset facilitates object reasoning and segmentation for visible and amodal masks across diverse scenes.

Along with the new task and dataset, we introduce a novel model named AURA (\textbf{A}modal \textbf{U}nderstanding and \textbf{R}easoning \textbf{A}ssistant). AURA can understand the user’s input questions, reason about the scene in the image, and predict both visible and amodal segmentation masks simultaneously. Given the challenge of predicting the shape of occluded parts, which are invisible in the image, we propose two key designs to enhance occlusion prediction capabilities. Specifically, we introduce the Occlusion Condition Encoder, which can distinguish whether objects are occluded and assess the occlusion degree comprehensively. Besides, the Spatial Occlusion Encoder can provide spatial guidance on whether each region is occluded, ensuring consistency between visible and amodal mask predictions.
It is worth noticing that the proposed AmodalReasonSeg dataset and the designed AURA method support reasoning and mask prediction for various objects within a single-turn conversation, enhancing applicability to real-world scenarios. Furthermore, AURA supports multi-round conversations, increasing its versatility.

Extensive experiments on the proposed AmodalReasonSeg dataset show that AURA outperforms existing methods in prediction accuracy, reasoning capability, and generalizability for complex occlusion scenarios.

The contributions of this paper can be summarized as follows: (1) We propose a novel task: amodal reasoning segmentation, along with a generalizable dataset collection pipeline. We construct a new dataset, AmodalReasonSeg, covering a variety of complex scenes from daily life. (2) We introduce AURA, a multi-modal method integrating vision and language for complex reasoning, capable of predicting both visible and amodal segmentation masks accurately. To the best of our knowledge, AURA is the first method explicitly designed for reasoning segmentation of amodal regions. (3) Extensive experiments demonstrate the effectiveness of AURA through both quantitative and qualitative evaluations.

\section{Related Work}

\subsection{Amodal Segmentation}

Amodal segmentation~\cite{li2016amodal} aims to predict the complete shape of an occluded object, encompassing both visible and occluded parts. Several methods~\cite{li2016amodal,qi2019amodal,follmann2019learning,li2023oaformer,liu2024plug,liu2024blade,li2023muva} are developed by extending the methods designed for visible segmentation or object detection~\cite{carion2020end,jiang2023modify,zhang2023black,jin2024llms,jiang2024weakly,jiang2024monomae}, while the relative occlusion order information is leveraged by many methods~\cite{zhang2019learning,zhang2024opnet,nguyen2021weakly,zhan2020self}. Since shape prior knowledge is an essential approach for inferring the shape of the occluded region, several methods~\cite{xiao2021amodal,li2023gin,li20222d,gao2023coarse,esser2021taming,sun2022amodal,chen2023amodal,ling2020variational} are designed to learn and utilize it. For instance, VRSP~\cite{xiao2021amodal} designs a codebook to store and retrieve shape priors for amodal masks. A3D~\cite{li20222d} learns to reconstruct 3D shape priors to guide amodal segmentation. C2F-Seg~\cite{gao2023coarse} works in a coarse-to-fine manner equipped with a Vector-Quantized codebook~\cite{esser2021taming}. 
Moreover, various datasets have been proposed for diverse scenarios, including daily life~\cite{zhu2017semantic,follmann2019learning}, indoor~\cite{zhan2024amodal,ehsani2018segan,dhamo2019object,zheng2021visiting}, autonomous driving~\cite{qi2019amodal,mohan2022amodal,reddy2022walt}, industrial~\cite{follmann2019learning}, shopping~\cite{li2023muva}, and gaming~\cite{hu2019sail}. 
However, existing amodal segmentation methods and datasets are limited to using images or videos as input, focusing solely on predicting amodal masks without user interaction or reasoning about implicit user purposes.
In this paper, we present the first high-quality dataset for the newly proposed amodal reasoning segmentation task and introduce a Vision-Language Model-based approach specifically designed for this task.

\begin{figure*}[htbp]
    \centering
    \includegraphics[width=1\linewidth]{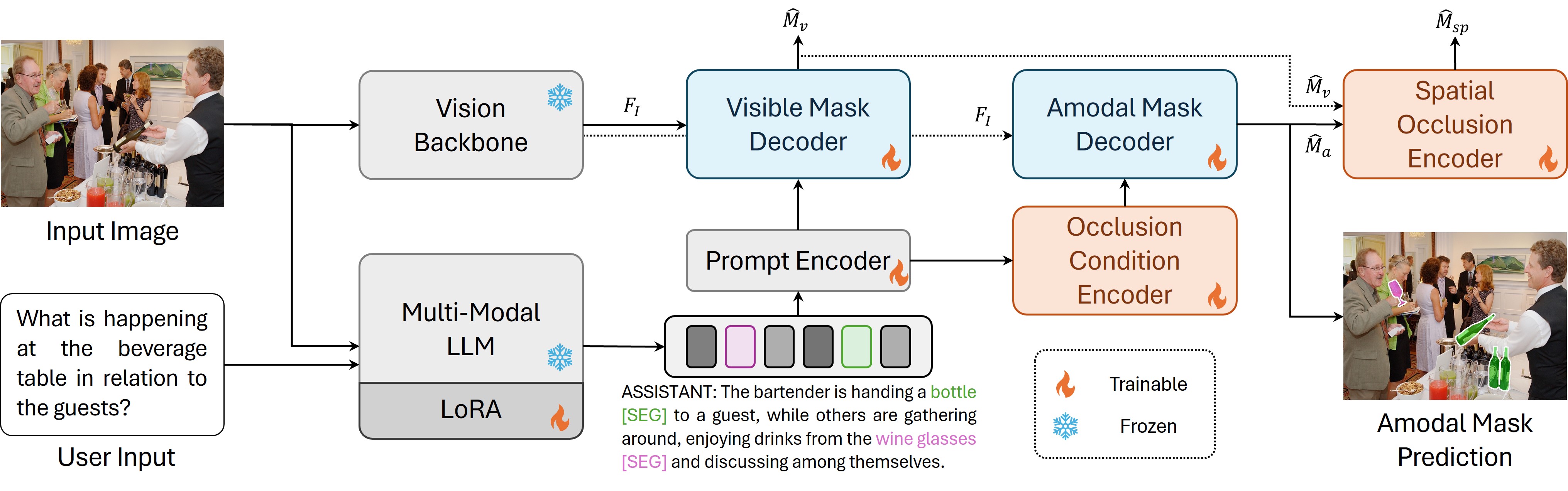}
    \caption{Overall architecture of the proposed AURA. (a) Given an input image and the input question from the user, the Vision Backbone extracts the visual features of the input image, and the Multi-Modal LLM equipped with LoRA is utilized for understanding the input image and textual questions simultaneously and responding with textual answers that include the [SEG] tokens indicating the segmentation masks. (b) For each [SEG], the Prompt Encoder takes its embedding of the Multi-Modal LLM and outputs the refined embeddings corresponding to the [SEG]. (c) Finally, the Visible Mask Decoder predicts the visible mask using each [SEG]'s refined embeddings. The Amodal Decoder predicts the amodal mask using the occlusion-aware embedding predicted by the Occlusion Condition Encoder. A Spatial Occlusion Encoder is designed to constrain the spatial occlusion information of the predicted visible and amodal segmentation masks to be accurate. Best viewed in color.}
    \label{fig:overall_architecture}
\end{figure*}

\subsection{Large Models for Reasoning Segmentation}

With the rapid advancement of Large Language Models (LLMs)~\cite{touvron2023llama,zhang2023llamaadapter,touvron2023llamav2} such as GPT-4~\cite{achiam2023gpt} and LLaMA~\cite{touvron2023llama}, language-based approaches have demonstrated impressive zero-shot generalization abilities across diverse real-world applications. Additionally, Vision Language Models (VLMs)~\cite{liu2024improved,liu2024visual,jiang2024multimodal} like BLIP-2~\cite{li2023blip} and LLaVA~\cite{liu2024visual} have achieved state-of-the-art performance in comprehensively understanding both visual and textual information. To leverage the reasoning capabilities of VLMs, LISA~\cite{lai2024lisa} introduces the task of reasoning segmentation, which predicts visible segmentation masks by reasoning based on user input questions. Building upon LISA, several recent methods, such as Pixel-LM~\cite{ren2024pixellm}, VISA~\cite{yan2024visa}, and LLM-Seg~\cite{wang2024llm}, have been proposed to enhance different aspects of the reasoning segmentation task, such as multi-object segmentation, video segmentation, and training efficiency, respectively. Despite these advancements, all existing methods are limited to segmenting only the visible region of target objects, leaving segmenting and analyzing the occluded region untouched. Our work aims to bridge this gap by enabling reasoning and segmenting in an amodal manner, specifically designed to handle complex occlusions in real-world scenarios.


\section{Methodology}
In this section, we first introduce the definition of the proposed amodal reasoning segmentation task in \ref{subsec:method_task_definition}. Then the overall architecture of the designed method is presented in \ref{subsec:method_overall_architecture}. Next, the special designs for learning and handling occlusion are introduced in \ref{subsec:method_occlusion_learning}. Finally, the loss functions that organize the training objectives are shown in \ref{subsec:method_loss_functions}.

\subsection{Task Definition}
\label{subsec:method_task_definition}

For the proposed amodal reasoning segmentation (ARS) task, each image $x_{img}$ of the dataset is associated with various questions $x_{text}$, answers $y_{text}$, and segmentations $y_{seg}$. For each pair, $x_{text}$ is a question from the user input that does not directly point to objects in the image. $y_{text}$ is the ground-truth textual answer that responds to the question and includes the description of one or more objects alongside the [SEG] token representing the position of the segmentation mask. $y_{seg}$ represents the ground-truth visible and amodal segmentation masks ($M_{v}$ and $M_{a}$), corresponding to the [SEG] in the answer $y_{text}$.

\subsection{Overall Architecture}
\label{subsec:method_overall_architecture}

As shown in Figure~\ref{fig:overall_architecture}, the overall architecture of the proposed AURA includes three stages. 
Specifically, (1) the Vision Backbone takes the input image $I$ as input for extracting its visual feature $F_{I}$, and the Multi-Modal LLM $\mathcal{E}_{MLLM}$ equipped with LoRA is employed to understand the input image $I$ and the user's input question $x_{text}$, and predict the textual answer $\hat{y}_{text}$ to respond the question. 
(2) The Prompt Encoder $\mathcal{E}_{P}$ takes the embedding $e_{mllm}$ of each [SEG] from the hidden states of the Multi-Modal LLM and predicts the refined embedding $e_{r}$ for each [SEG]. 
(3) Finally the $e_{r}$ and $F_{I}$ are fed into the Visible Mask Decoder $\mathcal{D}_{V}$ for predicting the visible segmentation mask $\hat{M}_{v}$ of each [SEG]. Besides, the $e_{r}$ is used by the Occlusion Condition Encoder $\mathcal{E}_{OC},$ which learns to predict the occlusion-aware embedding $e_{oa}$ that is contained within the occlusion condition information. Then the Amodal Mask Decoder $\mathcal{D}_{A}$ uses the $F_{I}$ and $e_{oa}$ for predicting the amodal segmentation mask $\hat{M}_{a}$ of each [SEG]. The Spatial Occlusion Encoder $\mathcal{E}_{SO}$ is designed to constrain the predicted $\hat{M}_{v}$ and $\hat{M}_{a}$ to conform to the spatial occlusion relationship, which implicitly improves the accuracy of the two predicted masks. 

The textual prediction of the Multi-Modal LLM is supervised using the $\mathcal{L}_{text}$, which can be denoted as:

\begin{equation}
    \mathcal{L}_{text} = \mathcal{L}_{CE}(\hat{y}_{text}, {y}_{text}),
\end{equation}

\noindent where $\mathcal{L}_{CE}$ is the Cross-Entropy loss~\cite{yi2004automated}. 
Moreover, for supervising the visible and amodal mask predictions, the $\mathcal{L}_{mask}$ is used as shown in the following:

\begin{equation}
\begin{split}
    \mathcal{L}_{mask} = \mathcal{L}_{CE}(\hat{M}_{v}, {M}_{v}) + \mathcal{L}_{CE}(\hat{M}_{a}, {M}_{a}) + \\
    \mathcal{L}_{Dice}(\hat{M}_{v}, {M}_{v}) + \mathcal{L}_{Dice}(\hat{M}_{a}, {M}_{a}),
\end{split}
\end{equation}

\noindent where $\mathcal{L}_{Dice}$ is the Dice loss~\cite{sudre2017generalised}. 

\subsection{Occlusion Learning}
\label{subsec:method_occlusion_learning}

\begin{figure}[t]
    \centering
    \includegraphics[width=0.92\linewidth]{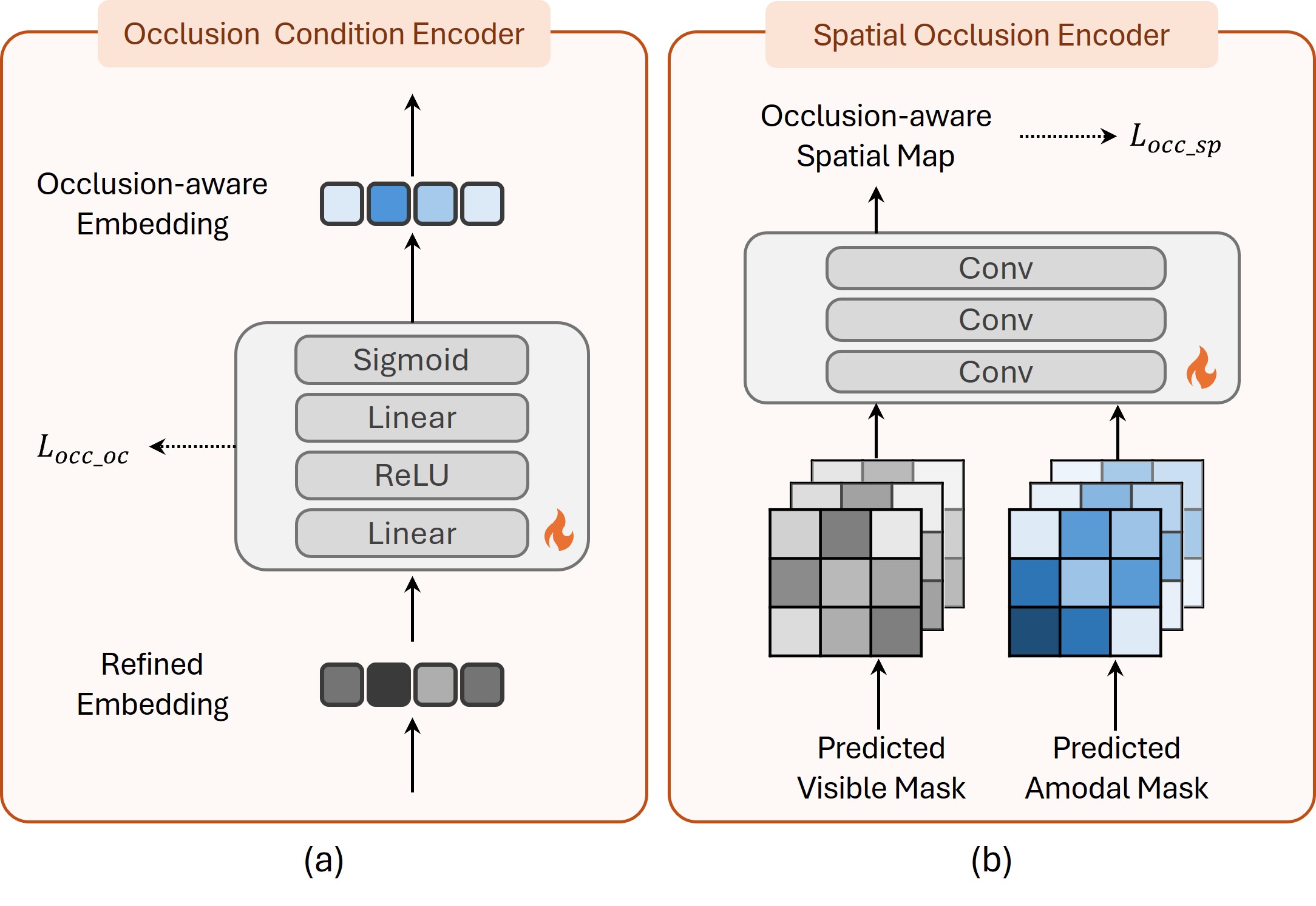}
    \caption{The designed occlusion encoders. (a) The Occlusion Condition Encoder learns to embed the occlusion condition of each mask and output the occlusion-aware embedding. (b) The Spatial Occlusion Encoder takes the predicted visible and amodal masks as input and predicts the occlusion-aware spatial map to constrain the spatial relationship between the two predicted masks.}
    \label{fig:occlusion_encoders}
\end{figure}

Figure~\ref{fig:occlusion_encoders} illustrates the designed occlusion encoders for handling occlusion learning in AURA. Detailed network structures, inputs and outputs, and losses are presented.

\noindent \textbf{Occlusion Condition Encoder} is designed to improve the ability of the refined embedding $e_{r}$ predicted by the Prompt Encoder to be aware of the occlusion condition of the object(s) corresponding to the [SEG]. Specifically, the $e_{r}$ is fed into the Occlusion Condition Encoder $\mathcal{E}_{OC}$ and predicts the occlusion-aware embedding $e_{oa}$ that is distinguishable by the degree of occlusion. The Occlusion Condition Encoder $\mathcal{E}_{OC}$ additionally predicts the occlusion rate $\hat{r}$ of the object, which reflects the occlusion condition comprehensively and is supervised by $\mathcal{L}_{occ-oc}$ with the ground-truth occlusion rate $r$. Moreover, a Linear layer and a Sigmoid layer are employed to predict the occlusion rate $\hat{r}$ based on the refined embedding $e_{r}$.

\noindent \textbf{Spatial Occlusion Encoder} is proposed to constrain the spatial occlusion conditions between the predicted visible masks $\hat{M}_{v}$ and amodal masks $\hat{M}_{a}$, enhancing the accuracy of visible and amodal segmentation prediction. Specifically, the Spatial Occlusion Encoder $\mathcal{E}_{SO}$ takes both $\hat{M}_{v}$ and $\hat{M}_{a}$ as input and predicts the occlusion-aware spatial map $\hat{M}_{sp}$, which is supervised by $\mathcal{L}_{occ-sp}$ with the ground-truth occlusion-aware spatial map ${M}_{sp}$. The ground truth ${M}_{sp}$ is defined by setting regions belonging to the ground-truth visible mask ${M}_{v}$ as 1 and setting regions belonging to the ground-truth amodal mask ${M}_{a}$ but not ${M}_{v}$ as 2, and setting other background regions as 0. In this approach, the constraint between predicted masks $\hat{M}_{v}$ and $\hat{M}_{a}$ is constructed and contributes to accurate segmentation.

For supervising the predictions of the occlusion condition and the spatial occlusion map, the $\mathcal{L}_{occ}$ is used as shown below:

\begin{equation}
    \mathcal{L}_{occ} = \mathcal{L}_{MSE}(\hat{r}, r) + \mathcal{L}_{CE}(\hat{M}_{sp}, {M}_{sp}),
\end{equation}

\noindent where $\mathcal{L}_{MSE}$ is the Mean Squared Error loss.

\subsection{Loss Functions}
\label{subsec:method_loss_functions}

The overall loss function $\mathcal{L}$ consists of four parts, supervising the textual answers $\hat{y}_{text}$ predicted by the Multi-Modal LLM, the predicted visible and amodal masks ($\hat{M}_{v}$ and $\hat{M}_{a}$), and the predictions of two designed occlusion encoders including the occlusion rate $\hat{r}$ and the occlusion-aware spatial map $\hat{M}_{sp}$. The overall loss function can be formulated as: 

\begin{equation}
    \mathcal{L} = \mathcal{L}_{text} + \mathcal{L}_{mask} + \mathcal{L}_{occ}.
\end{equation}

\section{AmodalReasonSeg Dataset}

Existing amodal segmentation datasets~\cite{qi2019amodal,follmann2019learning,zhan2024amodal} only contain images/videos and ground-truth masks without language annotations, which cannot fulfill the requirement of the amodal reasoning segmentation task. To this end, it is essential to construct a large-scale dataset with high-quality language annotations, including implicit user questions and the corresponding answers. 

In this section, we first introduce the basic information of the newly constructed dataset in \ref{subsec:overview_of_the_dataset}. Next, we demonstrate a general dataset generation pipeline that is generable for diverse scenarios and can be applied to generate new datasets efficiently in \ref{subsec:dataset_generation}.

\begin{figure}[t]
    \centering
    \includegraphics[width=0.9\linewidth]{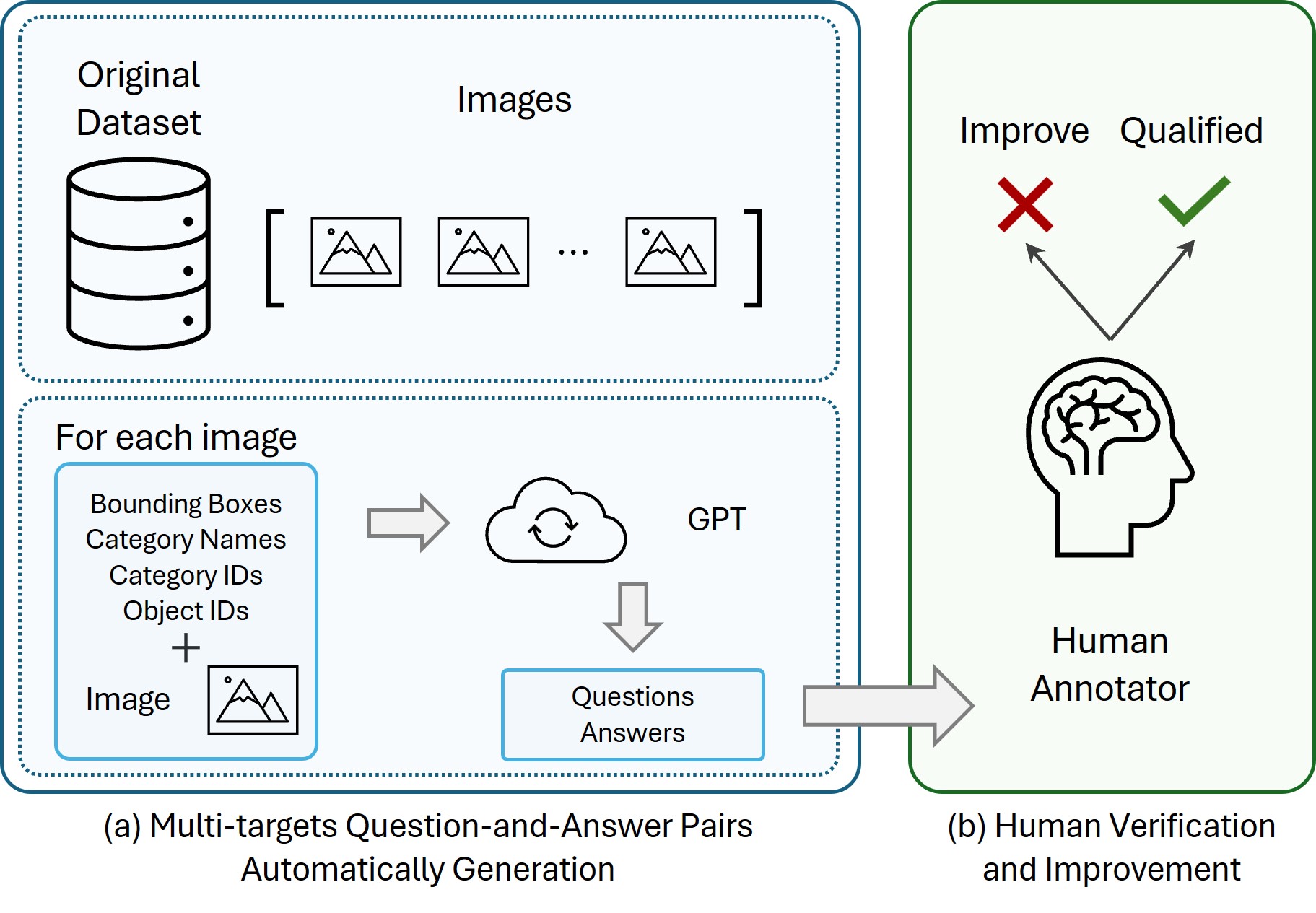}
    \caption{The designed generalizable pipeline for dataset generation. (a) Each image and the information of objects in it are gathered and fed into the GPT to generate questions and answers. (b) Then human annotators will check the generated data and improve it manually. Cross-check ensures the quality of the generated data.}
    \label{fig:pipeline_of_dataset_generation}
\end{figure}

\subsection{Overview of the Dataset}
\label{subsec:overview_of_the_dataset}

To construct the new dataset for the amodal reasoning segmentation task, the challenging amodal segmentation dataset COCOA-cls~\cite{follmann2019learning,zhu2017semantic} is utilized. The COCOA-cls dataset is built upon the famous COCO~\cite{lin2014microsoft} dataset, which includes daily scenarios with indoor and outdoor scenes.

For benchmarking the proposed amodal reasoning segmentation (ARS) task, we present the AmodalReasonSeg dataset, which is built upon the COCOA-cls dataset. The AmodalReasonSeg dataset includes 3,143 images and 35,494 high-quality question-and-answer pairs corresponding to the visible and amodal segmentation masks. The visible and amodal bounding boxes and the category information are included in our dataset.
For each image, there are 11.3 pairs of questions and answers on average annotated to cover diverse potential conversations. Overall, there are 2,018 images and 22,856 pairs of questions and answers in the train set and 1,125 images and 12,638 question-and-answer pairs in the test set.

Compared with existing amodal segmentation datasets, the AmodalReasonSeg dataset supports interacting with users by using textual questions as input, and the answers respond to the implicit purpose underneath the questions and alongside the visible and amodal segmentation predictions. 
Moreover, it is worth noticing that the answer corresponding to each image could refer to various objects, meaning the proposed dataset supports reasoning and predicting different targets in a single conversation, which is essential in real-world applications. 

\subsection{Dataset Generation}
\label{subsec:dataset_generation}

A generalizable pipeline is designed to generate the new dataset, as shown in Figure~\ref{fig:pipeline_of_dataset_generation}. Since manually annotating the question-and-answer pairs and the visible and amodal segmentation masks is labor-intensive and time-consuming, we present a semi-automatic pipeline inspired by~\cite{ren2024pixellm} to keep the efficiency of the annotating process and high-quality annotations. Specifically, the images and annotations, including category information and visible and amodal masks, are reserved from the original dataset. For generating the question-and-answer pairs, each image alongside the annotations is organized by a designed prompt template to be understood by a powerful and versatile large Vision-Language model, which is the ChatGPT-4o in our work that could take both image and text as input and output text response. 

The prompt template is carefully designed to require ChatGPT-4o to generate question-and-answer pairs that satisfy the requirements to ensure the generation of high-quality data. The annotation information of objects for this image is collected and fed to ChatGPT-4o. For each image, ten question-and-answer pairs are required to be generated. Different objects are required to be organized and contained in one answer. Besides, the relative occlusion information between different objects is required to be considered when generating the question-and-answer pairs, which makes the text description in the answer helpful for discriminating the occlusion relationship between objects.
The existing annotations from the original dataset of all objects in the image are organized and taken as input for the ChatGPT-4o. 
Finally, three human annotators check the generated questions and verify if they are qualified. During human verification, qualified samples are retained, while flawed ones are revised with appropriate questions and answers instead of being discarded. Common issues include incorrect object references, implausible functions, or inaccurate quantities. For the ones that need to be improved, human annotators will manually design appropriate questions and answers for these images. Cross-check is utilized between different human annotators to ensure the quality of the generated data.

\section{Experiments}

\subsection{Experimental Settings}

\noindent \textbf{Dataset.} All experiments, including our proposed method and compared methods, are trained on the training set of the proposed AmodalReasonSeg dataset, and evaluated on the validation set of this dataset for fairness. No extra dataset besides the AmodalReasonSeg dataset is used.

\noindent \textbf{Model Structure.}
We employ the LLaVA-7B-v1-1~\cite{liu2024visual} as the multi-modal LLM, which is equipped with LoRA for efficient fine-tuning. For the Vision Backbone and the Prompt Encoder, the SAM~\cite{kirillov2023segment} method with the ViT-H version is utilized. The mask decoder of SAM is employed for the Visible Mask Decoder and Amodal Mask Decoder.
Unless specifically specified, all experiments of the proposed AURA method are based on the model structure of using the LLaVA-7B-v1-1~\cite{liu2024visual} as the multi-modal LLM and the ViT-H version of SAM~\cite{kirillov2023segment}.

\noindent \textbf{Implementation Details.}
We implement our proposed AURA based on Pytorch~\cite{paszke2019pytorch}. The AdamW~\cite{loshchilov2017decoupled} optimizer is used with the learning rate set to 1e-3. The WarmupDecayLR learning rate adjustment policy is used with 100 iterations for warmup and then linearly decreases the learning rate to 0. 
Apart from the ablation studies using 13B models as Multi-Modal LLM Backbone that use two NVIDIA A100 GPUs (80GB Memory) for experiments, all of our experiments are conducted on two NVIDIA V100 GPUs (32GB Memory). The batch size for each GPU card is set to 1. The entire training iterations are 5000, and the gradient accumulation technique is used by setting the accumulation step to 4. The entire training time is about 10 hours.

\noindent \textbf{Evaluation Metrics.} The gIoU and cIoU metrics are used for evaluating the segmentation performance of amodal and visible masks. The gIoU is calculated by averaging the Intersection-over-Unions (IoUs) of all images, while the cIoU is calculated by dividing the cumulative intersection by the cumulative union of all masks.

\begin{table}[]
\centering
\setlength{\tabcolsep}{1.5mm}{
\scalebox{0.9}{
\begin{tabular}{lcccc}
\toprule
\multirow{2}{*}{Method} & \multicolumn{2}{c}{Amodal} & \multicolumn{2}{c}{Visible} \\ \cmidrule(l){2-5} 
         & gIoU & cIoU & gIoU & cIoU \\ \hline
\textit{Amodal Segmentation Methods} & & & \\ \hline
VRSP$^{\dag}$~\cite{xiao2021amodal}$_\text{~AAAI'21}$  & 15.18 & 16.55 & 17.41 & 19.75   \\ 
C2F-Seg$^{\dag}$~\cite{gao2023coarse}$_\text{~ICCV'23}$  & 18.62 & 18.25 & 19.12 & 20.31   \\ 
SD-Amodal$^{\dag}$~\cite{zhan2024amodal}$_\text{~CVPR'24}$  & 20.87 & 20.09 & 21.52 & 23.10  \\  \hline
\textit{MLLM-based Methods} & & & \\ \hline
LISA$^{\dag}$~\cite{lai2024lisa}$_\text{~CVPR'24}$     & 23.05 & 22.87 & 24.39 & 26.81  \\
LLM-Seg$^{\dag}$~\cite{wang2024llm}$_\text{~CVPR'24}$  & 24.62    & 23.06    & 25.19    & 27.94          \\
GSVA$^{\dag}$~\cite{xia2024gsva}$_\text{~CVPR'24}$      & 42.30    & 42.87    & 47.01    & 48.22    \\
OMG-LLaVA$^{\dag}$~\cite{zhang2024omg}$_\text{~NeurIPS'24}$ & 43.92    & 44.21    & 49.41    & 49.38        \\
GLaMM$^{\dag}$~\cite{rasheed2024glamm}$_\text{~CVPR'24}$     & 45.75    & 46.82    & 49.93    & 54.14     \\
PSALM$^{\dag}$~\cite{zhang2024psalm}$_\text{~ECCV'24}$     & \underline{46.81}    & \underline{47.03}    & \underline{50.64}    & \underline{55.19}    \\ 
\hline
Baseline & 43.55 & 43.92 & 48.96 & 48.63 \\
\rowcolor[HTML]{EFEFEF} 
AURA     & \textbf{47.76} & \textbf{47.32} & \textbf{51.31} & \textbf{55.38} \\ \bottomrule
\end{tabular}
}}
\caption{Comparison of the amodal reasoning segmentation performance on the AmodalReasonSeg dataset. $\dag$ denotes that this method is trained twice with ground-truth visible and amodal masks of the proposed AmodalReasonSeg dataset, respectively. Best in \textbf{bold}, second \underline{underlined}.}
\label{tab:reason_seg_results}
\end{table}

\subsection{Amodal Reasoning Segmentation}

\noindent \textbf{Compared Methods.} To benchmark the performance of amodal reasoning segmentation on the proposed AmodalReasonSeg dataset, state-of-the-art amodal segmentation methods~\cite{xiao2021amodal,gao2023coarse,zhan2024amodal} and multimodal large language model (MLLM)-based methods~\cite{lai2024lisa,wang2024llm,xia2024gsva,zhang2024omg,rasheed2024glamm,zhang2024psalm} for reasoning segmentation of visible region are used for comparison. All methods are trained and evaluated on the train set and the val set of our proposed AmodalReasonSeg dataset, ensuring fairness. LLaVA-7B-v1-1 is employed for all MLLM-based methods.

The amodal segmentation methods~\cite{xiao2021amodal,gao2023coarse,zhan2024amodal} are designed for taking an image as input and predicting amodal segmentation masks for all objects, but cannot understand the user's question, reasoning, and response with textual answers and segmentations.
Besides, MLLM-based methods~\cite{lai2024lisa,wang2024llm,xia2024gsva,zhang2024omg,rasheed2024glamm,zhang2024psalm} are designed to segment the visible regions of each object by taking the image and the user's question as input. 

However, these aforementioned methods can only predict either visible or amodal masks by training twice with ground-truth visible or amodal masks, while the proposed AURA can reason and predict the visible and amodal masks simultaneously. Moreover, MLLM-based methods like LISA can only reason for one object to each question, while the proposed AURA could reason and respond with various objects with textual answers describing them and corresponding segmentation predictions.
The baseline method is the proposed AURA without using the designed two occlusion learning modules, including the Occlusion Condition Encoder $\mathcal{E}_{OC}$ and the Spatial Occlusion Encoder $\mathcal{E}_{SO}$.

\noindent \textbf{Quantitative Results.} 
Table \ref{tab:reason_seg_results} shows the performance of all compared methods on the proposed AmodalReasonSeg dataset. $\dag$ means this method is trained twice on ground-truth visible and amodal masks, respectively, due to a lack of the ability to predict both masks at the same time. AURA achieves the best performance among all metrics regarding amodal and visible segmentations. To be more specific:
(1) Firstly, since amodal segmentation methods~\cite{xiao2021amodal,gao2023coarse,zhan2024amodal} are designed for amodal segmentation and cannot take a textual question as input and understand it, they cannot be certain of which objects to segment, and the predicted masks could not be for the target objects. Therefore, these methods are evaluated by computing metrics between all predicted objects and target objects.
In contrast, the proposed AURA can understand the purpose underneath the user's question and predict the amodal and visible masks for the targeted objects mentioned in the generated textual answer. 
(2) Secondly, since these MLLM-based methods are designed for segmenting visible regions of objects, their performance is not as well as AURA, which is specifically designed for the amodal reasoning segmentation task with task-specific designs for handling occlusions. Moreover, methods like LISA are designed to only segment one object for each question with other objects missed. However, AURA can support multi-object segmentation in a single conversation, responding with textual answers describing various objects with detailed elaborations.

\begin{figure*}[]
    \centering
    \includegraphics[width=0.86\linewidth]{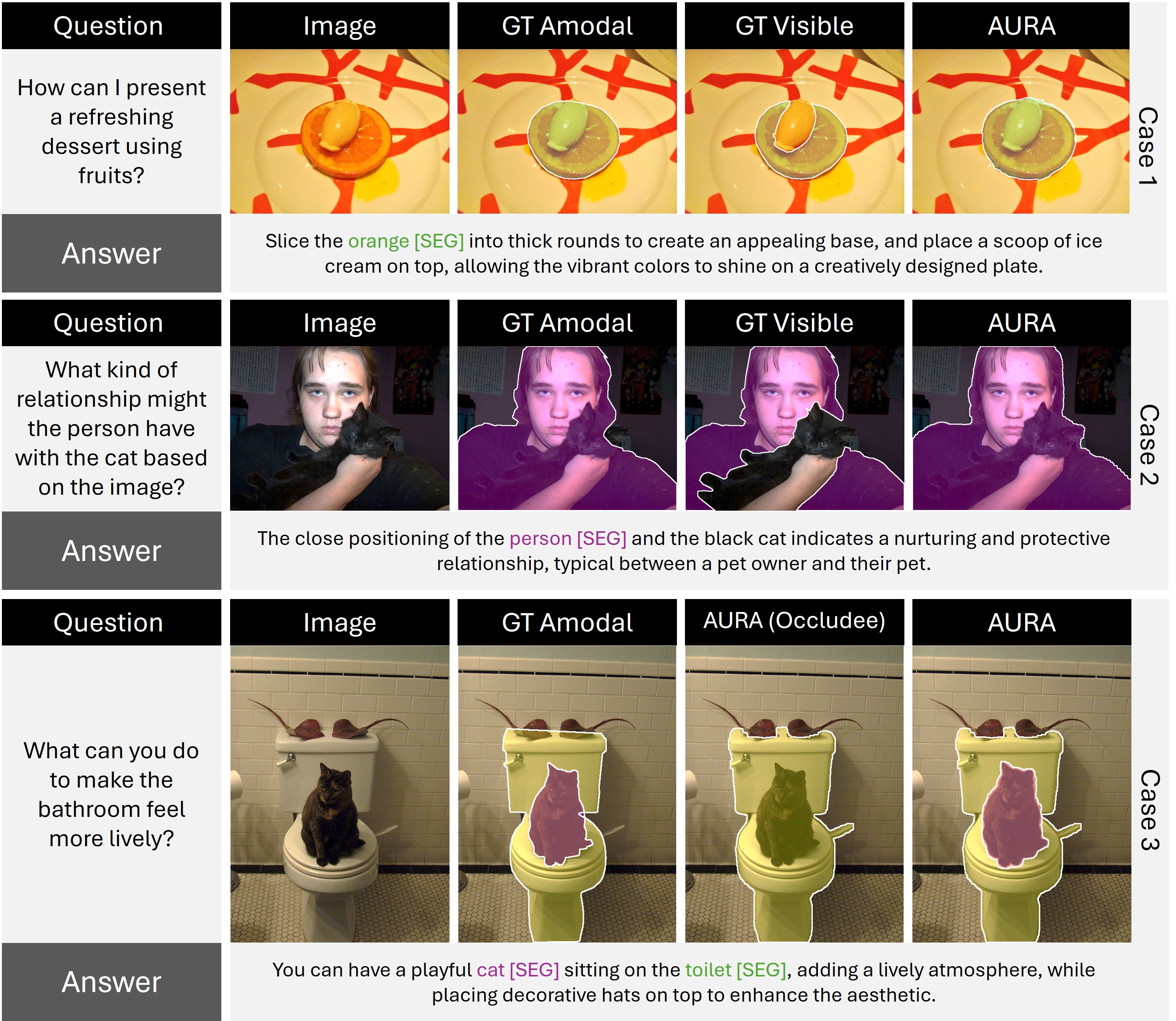}
    \caption{
    Visualization results of AURA. Three cases are shown from top to bottom with questions and ground-truth (GT) answers. The first two cases show the images, ground-truth amodal and visible masks, and the predicted masks of AURA. In the last case, the predicted amodal mask by AURA for the occludee, which is the occluded toilet, is shown replacing the ground-truth visible mask.
    }
    \label{fig:visualization_of_aura}
    
\end{figure*}

\noindent \textbf{Qualitative Results.}
Figure \ref{fig:visualization_of_aura} presents the visualization results of AURA in three cases, showing that AURA can accurately understand the occlusion relation between objects in the image, reason the purpose underneath the question, and predict segmentation masks for various objects. In the first case, although part of the orange is completely occluded by the ice cream on top, and the plate below has a messy pattern, which may cause confusion, AURA still can correctly understand the question and predict an intact amodal segmentation mask for the occluded orange. In the second case, the black cat and the person’s black clothing are easily confused, especially in a dark environment. However, AURA can precisely distinguish between the cat and the person's black clothing and accurately predict the complete amodal mask of the person, even though a large region of the person is occluded by the cat. In the third case, the front side of the toilet is occluded by the cat to a large extent, while AURA can reason the purpose of the question and predict a high-quality amodal segmentation mask for the occluded toilet and an accurate segmentation mask for the cat.

\subsection{Ablation Study}

We conduct ablation studies for the proposed AURA from various aspects, including the design of occlusion learning modules, the effectiveness of LoRA, the scale of the Multi-Modal LLM Backbone, and the scale of the Vision Backbone. 

\begin{table}[]
\centering
\setlength{\tabcolsep}{2.1mm}{
\scalebox{0.85}{
\begin{tabular}{ccccccc}
\toprule
\multicolumn{1}{c}{\multirow{2}{*}{Index}} &
  \multicolumn{1}{c}{\multirow{2}{*}{$\mathcal{E}_{OC}$}} &
  \multicolumn{1}{c}{\multirow{2}{*}{$\mathcal{E}_{SO}$}} &
  \multicolumn{2}{c}{Amodal} &
  \multicolumn{2}{c}{Visible} \\ \cmidrule(l){4-7} 
\multicolumn{1}{c}{} &
  \multicolumn{1}{c}{} &
  \multicolumn{1}{c}{} &
  \multicolumn{1}{c}{gIoU} &
  \multicolumn{1}{c}{cIoU} &
  \multicolumn{1}{c}{gIoU} &
  \multicolumn{1}{c}{cIoU} \\ \hline
1 &  &  & 43.55 & 43.92 & 48.96 & 48.63   \\			
2 & \checkmark & & 44.86 & 44.07 & 49.57 & 49.10   \\
3 & & \checkmark & \underline{46.95} & \underline{45.60} & \underline{50.68} & \underline{53.72} \\
4 & \checkmark & \checkmark & \textbf{47.76} & \textbf{47.32} & \textbf{51.31} & \textbf{55.38}  \\ \bottomrule
\end{tabular}}}
\caption{Ablation study of designs for occlusion learning. Best in \textbf{bold}, second \underline{underlined}.}
\label{tab:ablation_occ_learning}
\end{table}

\noindent \textbf{Occlusion Learning.} As shown in Table \ref{tab:ablation_occ_learning}, when the two designed Occlusion Condition Encoder $\mathcal{E}_{OC}$ and Spatial Occlusion Encoder $\mathcal{E}_{SO}$ are not used, the performance of AURA drops in a certain degree. When only using $\mathcal{E}_{OC}$, the occlusion conditions of all objects are learned and can help with our model for comprehensively distinguishing whether objects are occluded and the degree of occlusion for these objects. Besides, when only using $\mathcal{E}_{SO}$, the spatial occlusion distribution can be learned and guide our model to be aware of whether each region is occluded or visible, and constrain the consistency between the predicted visible and amodal masks regarding the spatial occlusion condition. Finally, when both $\mathcal{E}_{OC}$ and $\mathcal{E}_{SO}$ are utilized, the best performance is achieved with the learning of global and spatial occlusion conditions simultaneously.


\begin{table}[]
\centering
\setlength{\tabcolsep}{2.5mm}{
\scalebox{0.85}{
\begin{tabular}{lcccc}
\toprule
\multirow{2}{*}{Method} & \multicolumn{2}{c}{Amodal}                          & \multicolumn{2}{c}{Visible}                         \\ \cmidrule(l){2-5} 
                        & \multicolumn{1}{c}{gIoU} & \multicolumn{1}{c}{cIoU} & \multicolumn{1}{c}{gIoU} & \multicolumn{1}{c}{cIoU} \\ \hline
AURA w/o LoRA & 20.84 & 21.03 & 25.43 & 27.95 \\
AURA w/ LoRA  & \textbf{47.76} & \textbf{47.32} & \textbf{51.31} & \textbf{55.38}  \\ \bottomrule
\end{tabular}}}
\caption{Ablation study of the effectiveness of using LoRA in the Multi-Modal LLM for efficient fine-tuning. w/o and w/ denote without and with the LoRA technique, respectively. Best in \textbf{bold}.}
\label{tab:ablation_lora}
\end{table}

\noindent \textbf{Effectiveness of LoRA.} We conduct experiments to evaluate the effectiveness of LoRA, as shown in Table \ref{tab:ablation_lora}. The LoRA~\cite{hu2021lora} technique is utilized to be equipped with the Multi-Modal LLM for efficiently fine-tuning new data while keeping the pre-trained weight of LLM frozen. As shown in Table \ref{tab:ablation_lora}, when not using the LoRA technique, the performance of AURA drops to a certain extent. This is because the Multi-Modal LLM is not pre-trained on the AmodalReasonSeg dataset and has no specific knowledge learned to understand the purpose underneath the user's question and reason for various occluded objects. Therefore, the Multi-Modal LLM cannot predict textual answers correctly to respond to the user's question, leading to the performance of visible and amodal segmentation not well.


\begin{table}[]
\centering
\setlength{\tabcolsep}{2.15mm}{
\scalebox{0.87}{
\begin{tabular}{lcccc}
\toprule
\multirow{2}{*}{\makecell{Multi-Modal \\ LLM Backbone}} & \multicolumn{2}{c}{Amodal}                          & \multicolumn{2}{c}{Visible}                         \\ \cmidrule(l){2-5}  & \multicolumn{1}{c}{gIoU} & \multicolumn{1}{c}{cIoU} & \multicolumn{1}{c}{gIoU} & \multicolumn{1}{c}{cIoU} \\ \hline 
AURA-7B & 47.76 & 47.32 & 51.31 & 55.38  \\
AURA-13B  & \underline{49.41} & \underline{49.74} & \underline{54.13} & \underline{57.97} \\ 
AURA-Llama2-13B  & \textbf{52.34} & \textbf{52.95} & \textbf{58.42} & \textbf{60.21} \\ 
\bottomrule
\end{tabular}}}
\caption{Ablation study of the scale of the backbone for the Multi-Modal LLM. Best in \textbf{bold}, second \underline{underlined}.}
\label{tab:ablation_llm_backbone}
\end{table}


\begin{table}[]
\centering
\setlength{\tabcolsep}{3.2mm}{
\scalebox{0.85}{
\begin{tabular}{lcccc}
\toprule
\multirow{2}{*}{\makecell{Vision \\ Backbone}} & \multicolumn{2}{c}{Amodal}                          & \multicolumn{2}{c}{Visible}                         \\ \cmidrule(l){2-5}  & \multicolumn{1}{c}{gIoU} & \multicolumn{1}{c}{cIoU} & \multicolumn{1}{c}{gIoU} & \multicolumn{1}{c}{cIoU} \\ \hline 
SAM-ViT-B & 42.85 & 42.76 & 46.92 & 50.12 \\
SAM-ViT-L & \underline{45.23} & \underline{45.31} & \underline{49.18} & \underline{53.71} \\
SAM-ViT-H & \textbf{47.76} & \textbf{47.32} & \textbf{51.31} & \textbf{55.38}  \\ \bottomrule
\end{tabular}}}
\caption{Ablation study of the scale of the Vision Backbone. B, L, and H denote Base, Large, and Huge, respectively. Best in \textbf{bold}, second \underline{underlined}.}
\label{tab:ablation_vision_backbone}
\end{table}

\noindent \textbf{Scale of Multi-Modal LLM Backbone.} The effect of the scale of the Multi-Modal LLM backbone is evaluated as shown in Table \ref{tab:ablation_llm_backbone}. The AURA-7B, AURA-13B, and AURA-Llama2-13B denote the AURA method with different Multi-Modal LLM backbones corresponding to LLaVA-7B, LLaVA-13B, and LLaVA-Llama2-13B models. These models are different variants of the LLaVA model with different scales of parameters (7 billion and 13 billion) and different generations of Llama. All experiments, apart from this ablation study, are based on the AURA-7B model. We can observe from Table \ref{tab:ablation_llm_backbone} that the AURA-7B model does not perform as well as the other two models since the smaller number of parameters limits its learning ability for fine-tuning on the proposed AmodalReasonSeg dataset. Moreover, using larger models with 13 billion parameters can bring better performance since the learning capabilities of these models are better. Finally, upgrading the Llama large language model from the first to the second generation further enhances segmentation performance by generating higher-quality answers and refined embeddings used for segmentation.

\noindent \textbf{Scale of the Vision Backbone.} To evaluate the effect of the scale of the Vision Backbone, which is utilized to extract vision features of the input image and plays an essential role in providing informative features to the segmentation decoders, experiments are conducted, and the results are shown in Table \ref{tab:ablation_vision_backbone}. Three scales of the Vision Backbone are compared, including the Base, Large, and Huge versions of SAM, corresponding to 93.7M, 312M, and 641M parameters, respectively. We can observe that the model with a larger number of parameters can achieve better segmentation performance and greater learning ability.

\section{Conclusion}

In this work, we propose a new task named Amodal Reasoning Segmentation, aiming to predict textual answers alongside visible and amodal segmentation masks in response to the user's input question that requires understanding and complex reasoning in the image. A generalizable dataset generation pipeline is proposed, which can generate annotations with high quality efficiently. A new dataset named AmodalReasonSeg is presented for daily scenarios where occlusions are challenging for amodal segmentation. Extensive textual question-and-answer pairs are generated, and various objects are referred to in the answers. Moreover, we propose AURA to understand the user's implicit purpose and reason for the answer, which organizes various objects coherently with accurately predicted visible and amodal segmentation masks. Extensive experiments demonstrate the effectiveness of our proposed AURA on reasoning and segmenting the amodal masks in diverse complex scenarios. To enhance the applicability in real-world scenarios, we plan to further optimize the efficiency of our LLM-based model and enrich our dataset’s diversity and granularity. Moreover, feature compression approaches~\cite{hui2023rate,hui2024s,hui2025image,mi2024clif,mi2025mvqa} can be considered to improve the inference speed of the proposed approach.

\section*{Acknowledgments}

This research was supported by the Ministry of Education of Singapore Tier3 grant MOET32022-0006. This research was also supported by Culture, Sports and Tourism R\&D Program through the Korea Creative Content Agency grant funded by the Ministry of Culture, Sports and Tourism in 2024 (Project Name: Global Talent for Generative AI Copyright Infringement and Copyright Theft, Project Number: RS-2024-00398413, Contribution Rate: 10\%).